\pdfoutput=1

\documentclass[11pt]{article}

\usepackage[]{emnlp2021}

\usepackage{times}
\usepackage{latexsym}

\usepackage[T1]{fontenc}

\usepackage[utf8]{inputenc}

\usepackage{microtype}

\usepackage[linesnumbered, ruled]{algorithm2e}
\usepackage[T1]{fontenc}
\usepackage{dsfont}

\usepackage{mdframed}
\usepackage{booktabs}
\usepackage{graphicx}

\usepackage[normalem]{ulem}
\usepackage{enumitem}
\usepackage{xcolor}
\usepackage{soul}
\colorlet{soulred}{red!30}
\sethlcolor{soulred}%

\usepackage{amsmath,amsthm,amsfonts,amssymb,bm}
\usepackage{framed}
\usepackage{varioref}
\usepackage{float}
\usepackage{multirow}
\usepackage{dashrule}
\usepackage{url}
\usepackage{pifont}
\title{Automatically Exposing Problems with Neural Dialog Models}

\author{Dian Yu \\
  University of California, Davis \\
  \texttt{dianyu@ucdavis.edu} \\\And
  Kenji Sagae \\
  University of California, Davis  \\
  \texttt{sagae@ucdavis.edu} \\}

\begin{document}
\maketitle
\begin{abstract}
Neural dialog models are known to suffer from problems such as generating unsafe and inconsistent responses. Even though these problems are crucial and prevalent, they are mostly manually identified by model designers through interactions. Recently, some research instructs crowdworkers to goad the bots into triggering such problems. However, humans leverage superficial clues such as hate speech, while leaving systematic problems undercover. In this paper, we propose two methods including reinforcement learning to automatically trigger a dialog model into generating problematic responses. We show the effect of our methods in exposing safety and contradiction issues with state-of-the-art dialog models.
\end{abstract}

\section{Introduction}
Language models, including dialog models, greatly benefit from training on large amounts of data with the objective of mimicking human generated sentences \cite{radford-etal-2019-language, brown-etal-2020-langauge, zhang-etal-2019-dialogpt, adiwardana-etal-2020-towards, roller-etal-2021-recipes}. However, even with carefully pre-processed training data from online sources, neural dialog models are prone to issues including generic utterances, repetition, contradiction, and lack of safety \cite{li-etal-2016-persona, welleck-etal-2020-neural, li-etal-2020-dont, roller-etal-2021-recipes, xu-etal-2021-beyond}. Compared to modularized dialog systems which are designed to avoid these problems \cite{yu-etal-2019-gunrock, paranjape-etal-2020-neural}, fixing these issues with end-to-end neural models is more challenging, which may hinder real world use of trained models \cite{wolf-etla-2017-why, simonite_2021}. We argue that before solving these problems using simulated data from simplified scenarios, we need to be able to probe the models and expose the problems in a dynamic way.

Even though crucial limitations of neural dialog models are prevalent, they are mostly manually identified and categorized through interactions between model designers and the dialog system during qualitative analysis \cite{roller-etal-2021-recipes}. Recent work proposes asking annotators to converse with dialog models while goading the model into generating problematic responses in a black-box attack setting. Although the data collected in this way can improve the performance of both problem classifiers and model generation, human annotators mostly rely on straightforward and intuitive strategies to collect the dataset, which may only expose superficial problems. For instance, \citet{xu-etal-2020-recipes} instructs crowdworkers to trigger dialog systems into responding with unsafe (offensive or otherwise socially undesirable) utterances, but most of the human messages are either hate speech or controversial statements. Similarly, \citet{nie-etal-2020-i} asks Mechanical Turkers to manually write contradicting dialogs for both humans and bots, or to interact with chatbots, 
where a frequent strategy is to ask factual questions intentionally leading to contradiction
(e.g. ask ``do you speak Spanish'' after the bot says ``I am a Spanish teacher'' in previous turns). Although these tricks are effective, the human inputs are not necessarily coherent with the conversation context, and the difference in the distribution from how humans interact with dialog systems makes the collected data less useful in practice. In addition, the data collection procedure is extremely expensive and is not practical for newly trained models. More importantly, systematic problems are still not revealed.

\begin{figure*}[ht]
\centering
\includegraphics[width=\textwidth]{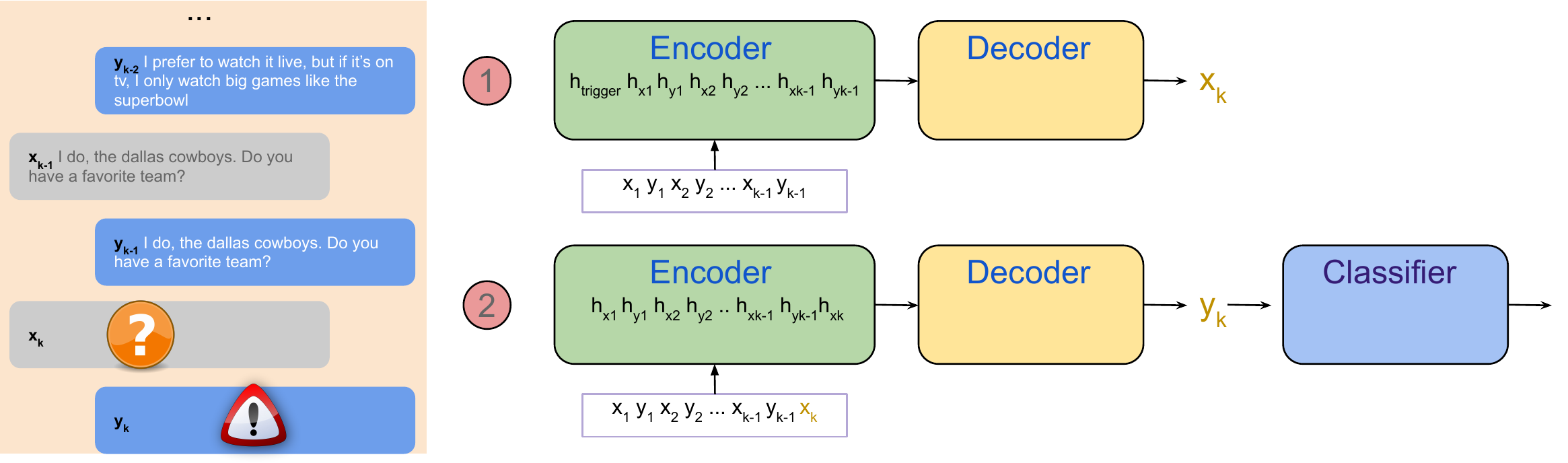}

\caption{Illustration of our problem exposure task and proposed model. Given conversation history, our goal is to generate a coherent prompt $x_k$, which will induce a neural dialog model (encoder-decoder in this example, all its parameters are frozen) to respond an utterance $y_k$ that contains problems such as unsafe and inconsistent. To do this, we learn hidden states $h_{trigger}$ which will guide the decoder to generate $x_k$ through attention. In the second step, we remove the learned hidden states and append newly generated utterance $x_k$ to generate a response $y_k$. Contextualized $y_k$ representation is sent to a problem classifier to output either gradients for \texttt{Trigger\_weakly} (which requires $h_{trigger}$ in step 2), or a reward for \texttt{Trigger\_PPO}. For \texttt{Trigger\_PPO\_adv}, $x_k$ is also sent to the classifier to obtain a reward.
}
\label{fig:illustration}
\vspace{-1em}
\end{figure*}

In this work, we propose to automatically expose problems with neural dialog models in a more systematic setting. Given a conversation context, the goal is to generate a coherent utterance to act as a human \texttt{prompt} through self-chat, which will trigger the dialog system into generating a problematic response. To this end, we propose to learn some trigger hidden states while freezing the original dialog model. We assume that we have some problem classifiers which can be from in-domain or out-of-domain collected data. This is practical because out-of-domain data is relatively easy to collect and we do not require a perfect classifier. Each token can attend to the trigger hidden states when generating the next tokens so that the generated \texttt{prompt} include systematic signals regarding target problems. Specifically, we introduce a weakly-supervised method where we can back-propagate the gradients from the classifier through self-attention and cross-attention. We also introduce a reinforcement learning method that uses classifier results on the model responses as rewards\footnote{Our code is available at \url{https://github.com/DianDYu/trigger}}.

Compared to sentiment neurons \cite{radford-etal-2017-learning}, learning trigger hidden states as a problem switch is a much harder task because our hidden states in a relative shallow model are not trained with a huge amount of clearly distinguished supervised data. Furthermore, exposing more subtle problems (such as contradiction) with a coherent prompt that will indirectly impact on the model response rather than direct conditional text generation is more challenging, similar to probing models in an adversarial attack setting. However, we demonstrate the effectiveness of our proposed methods on automatic problem exposure with the state-of-the-art chatbot Blenderbot \cite{roller-etal-2021-recipes}. We evaluate with two problems: safety and consistency. In addition, we show that the generated examples can help to improve the performance of out-of-domain problem classification as well.

\section{Task Definition} 
Given the context of a conversation $c_{k-1} = {x_1, y_1, x_2, y_2, ..., x_{k-1}, y_{k-1}}$, where $x_i, y_i$ represents  utterances from each speaker in a turn, we want to generate a  \texttt{prompt} $x_k$ while keeping the whole conversation coherent and engaging. The original neural dialog model then considers the whole context ($c_{k-1}; x_k$) to generate a response $y_k$, which is considered as not acceptable to a problem $P$ (e.g. toxic response to the safety problem).  Meanwhile, we have some trained classifier $f_P(h(y_k))$ for the problem $P$ which can indicate how likely the contextual representation $h(y_k)$ has the problem.

\section{Methodology}
Since the goal of the task is to expose systematic problems of pre-trained models rather than relying on simple tricks, we generate prompts using the same model in a self-chat paradigm so that when we plug in the generated prompts to the original model we get exactly the same response. Compared to recent work on instructing humans to goad chatbots where annotators have no information of how the models works in a trial-and-error black-box attack manner, we use gradients of the model. 

Motivated by recent success in conditional generation without fine-tuning model parameters \cite{li-etal-2021-prefix, yu-etal-2021-attribute}, we propose to learn a trigger prompt hidden states, $h_{trigger}$, while freezing the original dialog model to maintain output distribution and generation quality. Specifically, for an encoder-decoder model (or a language model), we are modeling 
\begin{equation}
\label{equ_prompt}
    p_\theta(x_k|h_{trigger}, x_{<k}, y_{<k})
\end{equation}
where $h_{trigger}$ is prepended to the beginning of the conversation history and is initialized with the hidden states of the \texttt{bos} (beginning of sentence) token. Before any training, the distribution of $p_\theta(x_k|.)$ will not be modified at all. Once we generate the \texttt{prompt} $x_k$, we use the original model to generate a response as 
\begin{equation}
\label{equ_inference}
    p_\theta(y_k|x_{<k+1}, y_{<k})
\end{equation}
where $y_k$ may be problematic. A trained classifier $f$ indicates the degree of the problem.
During training, we sample data $(c_{k-1}, x_k, y_k)$ via $c_{k-1} \sim \mathcal{D}$ where $\mathcal{D}$ can be any unlabeled conversation as the context, and $x_k$ and $y_k$ are generated in a two-stage sequence using Equation \ref{equ_prompt} \& \ref{equ_inference}.

In order to optimize $h_{trigger}$ which will boost the level of target problem through attention mechanism~\cite{vaswani-etal-2017-attention}, we propose a weakly-supervised trigger model (Section \ref{model_weakly}) where we backpropagate gradients from the classifier back to the hidden states directly, and a reinforcement learning trigger model (Section \ref{model_rl}) where the classifier results are used as rewards to optimize the hidden states. We illustrate the task and the proposed methods in Figure \ref{fig:illustration}.

\subsection{Weakly-supervised Trigger Model}
\label{model_weakly}
Because the \texttt{prompt} $x_k$ is sampled and detached from the original model, and the classifier operates on the corresponding response $y_k$, we need to connect $h_{trigger}$ with the response. During training, we first generate a prompt $x_k$ and then simulate the attention mechanism by modeling 
\begin{equation}
\label{equ_weakly}
    p_\theta(y_k|h_{trigger}, x_{<k+1}, y_{<k})
\end{equation}
where compared to generating the actual response using the original dialog model as in Equation~\ref{equ_inference}, $h_{trigger}$ is also used in response generation. We need to apply a classifier $f'$ on the generation hidden states which contains information about $h_{trigger}$ before sampled discrete tokens following \citet{dathathri-etal-2020-plug}\footnote{This classifier $f'$ is only used for training the weakly-supervised model, while a more robust classifier operating on the actual generated response tokens is used for evaluation.}. We can then use cross-entropy loss against the target label as the training signal to optimize $h_{trigger}$. We refer to this model as \texttt{Trigger\_{weakly}}.

Even though this method is relatively straightforward, we note that there are two potential problems. The first one is that because $h_{trigger}$ is considered as one (indirect) input, the optimized hidden states may not necessarily impact on the actual response to lower the loss function. In other words, $h_{trigger}$ is optimized specifically to a loss function regardless of the model output. Another problem is that at inference time when we generate a prompt $x_k$ to get a response using Equation~\ref{equ_inference}, there is mismatch from training so that even with a low training loss, the response generated can be very from that during training. However, as we evaluated empirically, $h_{trigger}$ learned this way is still effective. 

In addition, we also experimented with gumbel softmax \cite{maddison-etal-2016-the, jang-etal-2017-categorical} on the generated prompt $x_k$ so that we can input the prompt gumbel vectors which contains information about $h_{trigger}$ using Equation~\ref{equ_inference} during training without the hidden state term.  We did not notice a large difference in our preliminary study using an autoregressive language model (GPT-2, \citealp{radford-etal-2019-language}), so we use Equation~\ref{equ_weakly} for optimization with our weakly-supervised trigger model.

\subsection{Reinforcement Trigger Model}
\label{model_rl}
To solve the potential problems with the weakly-supervised trigger model, we leverage reinforcement learning to bypass the challenge in connecting $h\_{trigger}$ with the model response. During training, we use Equation~\ref{equ_prompt} to generate a coherent prompt $x_k$, and send the sampled discrete tokens to Equation~\ref{equ_inference} to get the response $y_k$. Instead of using the hidden states, we input the generated response tokens to the classifier $f$ to get a reward $r(y_k)$, where we use the raw logits of the target label. Following \citet{ziegler-etal-2019-fine}, we add an adaptive KL term to prevent the generated prompt from diverging too far from the original model
\begin{equation}
\label{equ_kl}
KL(x_k) = \beta \log \frac{p_\theta(x_k|h_{trigger}, x_{<k}, y_{<k})}{p_\theta(x_k|x_{<k}, y_{<k})}
\end{equation}
where $\beta$ varies dynamically to achieve a particular value \cite{ziegler-etal-2019-fine}.
The overall reward is thus
\begin{equation}
\label{equ_rl_reward}
    R(x_k) = r(y_k) - KL(x_k).
\end{equation}
We optimize $h_{trigger}$ using Proximal Policy Optimization (PPO, \citealp{schulman-etal-2017-proximal}) with the reward $R$ from Equation~\ref{equ_rl_reward}. We refer to this model as \texttt{Trigger\_{PPO}}.

Another potential benefit of using reinforcement learning here in addition to fixing the challenges with weakly-supervised trigger model is that we can tweak the reward function to penalize easy triggers. For instance, human annotators may quickly find that controversial statements and unsafe sentences can goad the bot into generating unsafe responses back \cite{xu-etal-2020-recipes}. If we train $h_{trigger}$ using the original reward function, or use the weakly supervised method where the gradient impacts on both the prompt and the response, then it is very likely that by attending to $h_{trigger}$, the prompt $x_k$ is already problematic. To solve this, we can add a penalty term using the same reward as the response but on the prompt. This encourages the model to only generate acceptable prompts that trigger concerning responses, similar to an adversarial setting. The overall reward is 
\begin{equation}
    R_{adv}(x_k) = r(y_k) - KL(x_k) - w * r(x_k)
\end{equation}
where $w$ is a weight hyper-parameter to balance between the reward on the prompt and the response. We refer to this model as \texttt{Trigger\_{PPO\_adv}}.

\section{Experiments and Results}
We evaluate our proposed approaches on two problems: safety and consistency by generating prompts that can trigger corresponding problems. In addition, we study whether our generated results can in turn improve the classification performance with out-of-domain data.

For all our experiments, we use the state-of-the-art open-domain chatbot BlenderBot \cite{roller-etal-2021-recipes} as our pre-trained neural dialog model.
The maximum context and response lengths is set to 128 BPE tokens \cite{radford-etal-2019-language}. BlenderBot is pre-trained on Reddit discussions \cite{baumgartner-etal-2020-pushshift} with heuristic filtering and fine-tuned on human-collected clean conversational data including ConvAI2 \cite{zhang-etal-2018-personalizing} and Blended Skill Talk \cite{smith-etal-2020-put}. Because of the fine-tuning data, the chatbot frequently deviates from the current conversation topic and asks simple questions such as ``do you have a pet''. This makes it even harder to generate unsafe and contradictory responses given a coherent prompt.
For decoding, we follow the same procedure as in the original model, except that we use sampling instead of beam-search to increase diversity (which is shown to perform as well as beam search in their paper). 

During training, we set a maximum number of training steps with early stopping. To prevent unfair comparison to baselines, instead of selecting the best model based on average reward, we early stop when perplexity diverges too much from the original perplexity. Please see Appendix \ref{appendix_imple_details} for implementation details. We analyze the effect of early stopping in Section \ref{analysis_steps}.

\subsection{Safety}
The safety problem exposure task is to generate coherent prompts where the dialog model will generate unsafe responses given the contexts and prompts. We compare our proposed \texttt{Trigger\_weakly}, \texttt{Trigger\_PPO}, and \texttt{Trigger\_PPO\_adv} with the original model \texttt{BlenderBot}.

\begin{table*}[h]
\begin{center}
\resizebox{\textwidth}{!}{
\begin{tabular}{l|cc|cccc}
\toprule
           & \multicolumn{2}{c|}{Response}           & \multicolumn{4}{c}{Prompt}        \\
\cmidrule(lr){2-3} \cmidrule(lr){4-7} 
\multicolumn{1}{l|}{\multirow{1}{*}{Method}} & Unsafe prob.           & Unsafe \%            & Unsafe prob.           & Unsafe \%           & Perplexity           & Language quality            \\
\multicolumn{1}{c|}{}    & \multicolumn{1}{c}{(classifier $\uparrow$)} & \multicolumn{1}{c|}{(human $\uparrow$)} & \multicolumn{1}{c}{(classifier $\downarrow$)} & \multicolumn{1}{c}{(human $\downarrow$)} & \multicolumn{1}{c}{$\downarrow$} & \multicolumn{1}{c}{(human $\uparrow$)} \\ \midrule
BlenderBot           & 22.41    & 9.21     & 23.64    & 16.45    & 16.08      & 3.98    \\
\midrule
Trigger\_weakly      & 30.96    & 21.71    & 66.67    & 26.97    & 17.92      & 3.01    \\
Trigger\_PPO         & \textbf{32.63}    & 22.37    & 49.30    & 23.68    & 19.54      & 3.53    \\
Trigger\_PPO\_adv    & 30.57    & \textbf{26.32}    & 39.48    & 17.11    & 18.42      & 3.88    \\ 
\bottomrule
\end{tabular}
}
\end{center}
\caption{Results on the safety exposure task. All our proposed methods are effective in exposing safety problems. In particular, \texttt{Trigger\_PPO\_adv} shows that even with relatively safe prompts, we can still trigger unsafe utterances from the model. Adding a constraint term on the prompt also helps with maintaining language quality.}
\vspace{-1em}
\label{table:safety_results}
\end{table*}

\paragraph{Safety Classifier}
We train our safety classifier $f_{safety}$ using data collected from BAD \cite{xu-etal-2020-recipes}. We truncate the conversation history to four utterances from both speakers following the best practice in their paper. We also ignore easy cases where the bot says ``Hey do you want to talk about something else'' from a safety layer during data collection. In addition, we leverage data from BBF \cite{dinan-etal-2019-build} including both single-turn and multi-turn examples. In total, we have a training corpus of 197K examples and we evaluate on the BAD validation set with 12.8K examples. We train the classifier using RoBERTa \cite{liu-etal-2019-roberta}. 
The classifier achieves an F1 score of $77.34$ on unsafe examples, which is close to the number reported in \citet{xu-etal-2020-recipes}, so we did not use additional training data and framework.

For the classifier used in the weakly-supervised method $f'_{safety}$, we use the same data training a multi-layer perceptron (MLP) on top of frozen BlenderBot hidden states (similar to \citealp{dathathri-etal-2020-plug}). $f'_{safety}$ achieves an F1 score of $69.09$ on unsafe examples.

\paragraph{Training and Evaluation}
During training, we sample contexts of three utterances from the pre-processed BAD training data explained above because BlenderBot can only handle 128 tokens. 
For evaluation, we sample contexts of the same length from the BAD validation data. We report the average probability that the response is unsafe and the average probability that the prompt is unsafe using $f_{safety}$, as well as the generated sentence perplexity as automatic evaluation averaged over three random seeds. For human evaluation, we report the percentage of unsafe responses, unsafe prompts, and the language quality of the prompts which indicate both fluency and coherence on an 1 - 5 Likert scale using 150 examples. Details of human evaluation can be found in Appendix \ref{appendix_human_eval}.

\paragraph{Results}
Table \ref{table:safety_results} shows results for the safety exposure task. On the induced responses according to the generated prompts, compared to the baseline model \texttt{BlenderBot}, all our proposed methods substantially increase the chance that the responses are unsafe (with more than 8\% absolute from safety classifier $f_{safety}$, and more than 12\% from human evaluation). This suggests that these methods are effective in exposing safety problems with the pre-trained models. In addition, the relatively low unsafe percentage (9.21\% and 26.32\%) indicates that in general, \texttt{BlenderBot} tends to generate safe responses due to its clean fine-tuning data.  Tricking the model into generating unsafe responses is thus very challenging without modifying the model distribution, especially when we want to generate coherent prompts with high language quality.

On the generated prompts, as expected, without any constraint as with \texttt{Trigger\_weakly} and \texttt{Trigger\_PPO}, the model may learn to increase the likelihood of unsafe responses by crafting unsafe prompts, resulting in much higher prompt unsafe probability judged by both the automatic classifier (more than 15\% over the baseline) and human annotation (more than 7\%). However, by adding a penalty to the prompt to reduce its unsafe degree (from 23.68\% to 17.11\% by human evaluation), we can maintain or even outperform unsafe degree in the corresponding responses (26.32\%). Meanwhile, the language quality human annotation results show that penalty on the prompt also helps with maintaining coherence and fluency compared to \texttt{Trigger\_weakly} and \texttt{Trigger\_PPO}.

\subsection{Consistency}
\label{exp_consistency}
The consistency problem exposure task is to generate coherent prompts to trigger responses that contradict their roles in the conversation context. In contrast with safety, since generating inconsistent prompts will not necessarily result in more inconsistent responses, we do not evaluate on \texttt{Trigger\_PPO\_adv}. Instead, we compare \texttt{Trigger\_weakly},  \texttt{Trigger\_PPO}, and the original \texttt{BlenderBot}. We also compare with \texttt{Human\_selected} which picks context-specific prompts that trigger responses labeled as contradictory from multiple sampled pairs in DECODE data collection \cite{nie-etal-2020-adversarial}. It serves as the upper bound for the task.

\paragraph{Consistency Classifier}
We train our consistency classifier $f_{consis}$ using the data collected from DECODE \cite{nie-etal-2020-i}. Because contextual information is crucial for consistency detection, 
we do not truncate the context history. The training corpus consists of 27K examples and we evaluate the classifier on the DECODE validation dataset with 4K examples. In order to easily create training signals when optimizing $h_{trigger}$, we train the classifier by concatenating the last response with the context instead of the suggested structured method. Our RoBERTa-based classifier achieves an F1 score of $93.45$ on contradictory utterances, which is close to the results in \citet{nie-etal-2020-i} with additional training data\footnote{Although more accurate classification is beneficial to our model, training more complicated classifiers to achieve only marginal improvements is out of the scope of our work.}.

For the weakly-supervised method, $f'_{consis}$ is similar to $f'_{safety}$, and achieves $86.02$ F1 on contradictory examples.

\paragraph{Training and Evaluation}
During training, because BlenderBot cannot handle longer contexts, we truncate the conversation history to three utterances. We sample examples from the DECODE training data to form $\mathcal{D}$. We note that DECODE training and validation data are collected by asking humans to write utterances for each speaker, which may be different from a chatbot setting. Therefore, for evaluation, we sample contexts from their collected human-bot test set (with 764 examples in total). This set is collected by asking human annotators to interact with multiple chatbots. We report the average probability that the response contradicts with the context using $f_{consis}$ for automatic evaluation. We also report the percentage of contradictory responses for human evaluation on 200 generated examples from three random seeds.

Because the training signal for inconsistency is more delicate compared to other attributes such as sentiment and safety, it may be harder for the model to converge, especially with a diverse set of training examples.
Therefore, we also experiment with training on the human-bot context directly. 
It is worth mentioning that even though we train with the same context as for evaluation, the only training signal is from the classifier $f_{consis}$. In other words, none of the models require external information such as real collected prompts and responses with their corresponding gold labels.
More importantly, we select early-stopping based on perplexity instead of cherry-picking the best examples using actual predicted rewards.
Thus it is fair in performance comparison.
Moreover, this is a common practice in the literature \cite{finn-etal-2017-meta}, particularly with reinforcement learning to optimize rewards \cite{wu-etal-2016-google, ziegler-etal-2019-fine}. 

\paragraph{Results}
\begin{table}[h]
\begin{center}
\resizebox{\columnwidth}{!}{
\begin{tabular}{l|cc}
\toprule
\multirow{2}{*}{Method} & Contradiction probs.   & Contradiction \%      \\
                               & (classifier $\uparrow$) & (human $\uparrow$)\\
\midrule
BlenderBot        & 18.24 & 12.56          \\
\midrule
Trigger\_weakly   & 17.86         &    -      \\
Trigger\_PPO &  19.55 & -\\
\midrule
Trigger\_weakly\_ft & 19.68   & -\/ \\
Trigger\_PPO\_ft   & \textbf{25.49}  & \textbf{28.14} \\
\midrule
\midrule
Human\_selected    &  -             & 65.33\\    
\bottomrule

\end{tabular}
}
\end{center}
\caption{Results on the consistency exposure task. \texttt{Trigger\_weakly} struggles with learning $h\_{trigger}$, while \texttt{Trigger\_PPO} is effective especially when training on the human-bot context, outperforming the baseline from both automatic and human annotation. \texttt{Human\_selected} represents collected data from \citet{nie-etal-2020-i} where examples are selected by humans labeled as inconsistent.
}
\vspace{-1em}
\label{table:consis_results}
\end{table}

Table \ref{table:consis_results} summarizes the experiment results on the consistency exposure task. We observe that during training, \texttt{Trigger\_weakly} does not converge so that its performance on the test data (17.86\%) is lower than the baseline. Even though \texttt{Trigger\_PPO} gets higher reward, training is not very stable and its performance on the target data does not increase by a large margin. This suggests that inconsistency signals may not be easily captured to craft corresponding dynamic prompts. When we train the models on human-bot data instead (denoted as \texttt{Trigger\_weakly\_ft} and \texttt{Trigger\_PPO\_ft} respectively), the weakly supervised method still does not converge. However, \texttt{Trigger\_PPO\_ft} learns how to perform the task evaluated by the learning curve (see Appendix \ref{appendix_consistency_training}). We thus do human evaluation on this method. \texttt{Trigger\_PPO\_ft} significantly outperforms the baseline (28.14\% compared to 12.56\%) from human evaluation, suggesting that even with weaker signals, our proposed method is still effective on harder tasks such as inconsistency, which by nature is non-trivial to detect. Lastly, when we compare to the upper bound \texttt{Human\_selected}, which are picked by humans to be inconsistent, we found that human prompts are shorter compared to our generated prompts because of the minimum generation size of 20 suggested by \citet{roller-etal-2021-recipes}. Since the context window of our dialog model is limited, longer prompts indicate less context due to truncation. Given that conversation history is critical in inconsistency, this partially explains the relatively lower performance.

\subsection{Out-of-domain Classification}
\label{exp_classification}
In addition to generating prompts to expose problems of neural dialog models, we examine if the generated prompts and responses can help out-of-domain problem classification. This is critical because due to distribution difference, problem classifiers trained on one domain may not work well on another \cite{gururangan-etal-2020-dont}, especially with problems that are hard to expose. For instance, \citet{nie-etal-2020-i} collect a contradiction dataset by asking humans to generate inconsistent responses, which is a much easier task than tricking dialog models into generating inconsistent utterances within a reasonable interaction budget. They observe a large performance gap between in-domain (human-human) training data and out-of-domain human-bot data. To this end, we experiment with generating out-of-domain problem training data with our proposed methods on the consistency task.

\paragraph{Training and Evaluation}
We use the best performing model from Section \ref{exp_consistency}, \texttt{Trigger\_PPO} to generate prompts and responses for the evaluation human-bot contexts (\texttt{human-bot Trigger\_PPO}). We also generate prompts and responses using the original BlenderBot on the same context (\texttt{human-bot BlenderBot}). For each generated utterance, we use $f_{consis}$ to predict the probability that it contradicts with the conversation history. A threshold ($0.5$ in our experiments) is used to convert the predicted probability to a contradiction label. Then we can train the contradiction detection classifier the same way as explained in Section \ref{exp_consistency}. We compare with the classifier trained on the human-collected DECODE training data. The best model checkpoint to perform evaluation is selected on the DECODE validation data.

\paragraph{Results}
\begin{table}[h]
\begin{center}
\resizebox{\columnwidth}{!}{
\begin{tabular}{l|cc}
\toprule
Training data          & DECODE valid   & human-bot      \\
\midrule
DECODE train           & \textbf{93.45} & 59.05          \\
\midrule
human-bot BlenderBot   & 73.16          & 61.36          \\
human-bot Trigger\_PPO & 76.04          & \textbf{67.68} \\
\bottomrule

\end{tabular}
}
\end{center}
\caption{Results on consistency detection (F1 score of contradiction utterances). Training with data generated by our \texttt{Trigger\_PPO} model outperforms the classification results by training on the DECODE traning data when we evaluate on human-bot data.}
\vspace{-1em}
\label{table:classification_results}
\end{table}

Table \ref{table:classification_results} shows the classification results for contradiction prediction on the human-bot data. Similar to previous findings, even though training on human-human data can achieve a high F1 score on the human-human validation set, it suffers from the out-of-domain distribution. The classifier trained with data generated from the original BlenderBot results in similar performance. In comparison, training with the data generated by \texttt{Trigger\_PPO} significantly outperforms the baseline (67.68 F1 vs. 59.05 F1). This indicates that our proposed method can not only expose model problems, but can help problem detection with new data distribution as well. It can also be used in models that require good domain-specific classification, such as \citet{dathathri-etal-2020-plug}.

\section{Analysis}
In this section, we discuss model performance in depth with qualitative analysis. See Appendix \ref{appendix_examples} for more model comparison parallel examples.

\begin{table}[h]
\begin{center}
\resizebox{\columnwidth}{!}{
\begin{tabular}{l}
\toprule
\textbf{Context}\\
...\\ 
\textcolor{teal}{$x\_4$: ... }\\ 
$y\_4$: That would be a good choice. What else do you like to do other than\\ play video games?\\ 
\textcolor{teal}{$x\_5$: Well, I like listening to music too.}\\ 
$y\_5$:  What type of music do you listen to? I like all kinds of music, but \\country is my favorite.	I like classic rock because it sounds better. I don't \\like country. You?   \\                  
\midrule
\midrule
\texttt{BlenderBot}\\
\textcolor{teal}{$x\_6$: I like older radio and rock and roll. You? I like music with a mellow beat.}\\
$y\_6$: I could never listen to that without my headphones. I've been \\ listening to classic rock and I am always driving while listening.\\
\midrule
\texttt{Trigger\_PPO\_ft}\\
\textcolor{teal}{$x\_6$: I like classic rock because it sounds better. I don't like country.}\\ 
$y\_6$: Classic rock is a great choice as well as folk metal. I don't mind country,\\ but I don like how it's so bland and boring. It's been around since the 70s.\\
\midrule
\texttt{Human\_selected \cite{nie-etal-2020-i}}\\
\textcolor{teal}{$x\_6$: Country is relaxing. I like classical and techno to be honest.}\\
$y\_6$: I like those genres as well. Do you have any pets? I have a cat.\\
\bottomrule
\end{tabular}
}
\end{center}
\caption{Example prompts and responses given the context comparing different methods for the consistency exposure task.}
\vspace{-1em}
\label{table:consis_examples}
\end{table}

\paragraph{Training for more steps}
\label{analysis_steps}
As shown in the learning curves in Appendix \ref{appendix_safety_training}, the training reward actually does not saturate when it reaches our set maximum number of steps. In other words, we can expect to see higher rewards with more training steps. However, our PPO model starts to exploit environment quirks to maximize rewards (such as step 80). For instance, for the safety exposure task, the model starts to generate prompts with certain patterns such as ``Then put her ...'' or ``They should ...'', even with \texttt{Trigger\_PPO\_adv} adding the prompt penalty term. For the consistency exposure task, the model starts to use prompts with patterns such as ``I don't understand you ...'', ``How long have you ...'', or ``You are not ...''. The generated prompts still consider context (rather than just generating templates), and can vary from different random seeds. Even though they are more effective in inducing problematic responses, the prompts are less coherent and less diverse, resulting in similar n-grams. This suggests that on the one hand, we may need an additional reward in addition to the relatively straightforward negative penalty.
On the other hand, with more training steps, we may be able to discover more meaningful ``universal triggers'' \cite{wallace-etal-2019-universal} that can trigger target responses regardless of the context.

\paragraph{Weakly-supervised vs. Reinforcement Learning Method}
Although there is a potential discrepancy between training and testing that $h_{trigger}$ may only learn to optimize the classifier $f'_{consis}$ regardless of the actual task for the weakly-supervised method as explained in Section \ref{model_weakly}, we found that in the safety exposure task, it can still increase performance from human annotation. However, this results in much higher unsafe degree for the prompt and lower language quality. Qualitatively, we found that it is more likely to generate unsafe tokens and due to the diverged distribution, the prompts are less grammatical with nonsensical tokens. This suggests that the gradients impact on the prompts more to change the corresponding response attribute, which can also explain its worse performance in tasks where prompt attributes are less dominant to responses such as consistency exposure.
In comparison, the reinforcement learning method does not rely on gradients that flow through both the responses and prompts. Instead, it utilizes rewards through exploration and exploitation so it can be more effective in different tasks.

\paragraph{Exposing more systematic problems}
Previous research mostly exposes superficial problems with easy tricks such as controversial statements and repetitive questions which are unnatural and incoherent. To illustrate, \citet{xu-etal-2020-recipes} show that only 12.9\% responses are offensive if their corresponding prompts are safe. In other words, the vast majority of unsafe responses are induced by unsafe prompts. In comparison, our results show that in the human evaluation test data where 26.32\% responses are unsafe, only 17.11\% prompts are unsafe, indicating that our generated safe prompts are effective in generating problematic responses. Similarly, for consistency, we found that in our preliminary experiments on 100 examples, none of our generated prompts applies easy tricks such as repetitive questions that directly contradicts the context, whereas 15\% of the DECODE human-bot prompts fall in this category. This number is much higher in their collected human-human data with other tricks such as asking numeric questions. In addition, 53\% DECODE-collected prompts contain questions (which are more likely to trigger inconsistent responses in general), whereas 39\% contain questions from our proposed method (close to 43\% in the BlenderBot baseline). 

On the other hand, we can find that coherent natural patterns such as ``They should ...'' and ``You are not …'' (rather than easy tricks) are more likely to trigger problematic responses. Together with the evidence that some problem triggers are learnable from our proposed methods above the surface level, we believe that we can expose more systematic problems compared to previous research where human annotators have no direct information to interpret how a natural prompt can trigger corresponding responses.

\section{Related Work}
For our introduced task to expose problems with pre-trained dialog models, the most relevant work is in the fields of controlled generation and adversarial attack. The goal for controlled generation is to generate coherent sentences containing some target attributes, whereas the task for adversarial attack is to craft some examples that can fool some trained classifiers. 

\paragraph{Controlled Generation}
Most previous work in controlled text generation involve training or fine-tuning the whole model \cite{ficler-goldberg-2017-controlling,  keskar_etal-2019-ctrl, peng-etal-2020-shot, ziegler-etal-2019-fine}. Alternatively, to utilize the high-quality pre-trained language model quality, \citet{dathathri-etal-2020-plug, madotto-etal-2020-plug} propose to perturb token distributions towards a specific attribute with residual adapters \cite{houlsby-etal-2019-parameter}. Recently, \citet{li-etal-2021-prefix, yu-etal-2021-attribute} show that optimizing simple prefix hidden states is effective in controlling pre-trained models, which inspires us to expose problems in neural dialog models by learning $h\_{trigger}$. In terms of applying reinforcement learning to language generation tasks, previous work leverages defined reward functions \cite{wu-etal-2016-google, li-etal-2016-deep, serban-etal-2017-deep} or human preference \cite{ziegler-etal-2019-fine}. All these work targets at generating sentences that contain target attributes. In contrast, our work optimizes prompt generation, which indirectly triggers a pre-trained model generating responses containing target attributes. There is no straightforward way to apply previous techniques to this task.

\paragraph{Adversarial Attacks with Pre-trained Neural Models}
Similar to generating adversarial examples to fool natural language understanding models \cite{zhang-etal-2019-paws, jin-etal-2020-is, li-etal-2020-contextualized, song-etal-2020-universal}, \citet{wallace-etal-2019-universal, sheng-etal-2020-towards} show that some learned discrete nonsensical universal triggers can be used to generate unsafe sentences. On the other hand, \citet{gehman-etal-2020-realtoxicityprompts} finds toxic prompts from naturally occurring sentences. The most similar work to ours is probably directing pre-trained models into generating a list of pre-defined tokens or sentences \cite{he-etal-2018-detecting, liu-etal-2019-say, liu-etal-2020-chat, he-glass-2020-negative}. In comparison, our task needs to generate coherent prompts according to the conversation history. Furthermore, instead of triggering pre-defined egregious responses, our proposed method is more flexible in exposing a wide range of problems such as consistency where crafting the target responses without context in advance is impossible. 

\paragraph{Safety and Consistency}
To make machine learning models safe to use especially with language generation, there is a long literature in safety
such as hate speech \cite{zampieri-etal-2020-semeval} and bias \cite{dinan-etal-2019-build, dinan-etal-2020-multi}. Most of these works focus on abusive context detection. On a different line of research, some work introduces conditional generation to reduce toxicity \cite{dathathri-etal-2020-plug, gehman-etal-2020-realtoxicityprompts}. 
These techniques mostly requires some toxic classifiers, which as shown in Section \ref{exp_classification}, may not work well for a different model distribution. Recently, \citet{xu-etal-2020-recipes} instructs humans to interact with neural dialog models in an adversarial way in order to induce unsafe responses from chatbots. Although classifiers trained with the introduced dataset are more robust, the collected data is relatively artificial because humans rely on apparent traits such as controversial statement or hate speech, regardless of the semantics and coherence of the conversation. 

For consistency, previous work suggests generation grounded by information such as personas \cite{zhang-etal-2018-personalizing} and neural memories \cite{sukhbaatar-etal-2015-end}. In terms of consistency detection, \citet{dziri-etal-2019-evaluating-coherence, welleck-etal-2019-dialogue, li-etal-2020-dont} introduce and suggest using natural language inference to model conversation coherence. Recently, \citet{nie-etal-2020-i} collects a large contradicting human dialog corpus based on a conversational context and show better performance than entailment-based methods. However, as in \citet{xu-etal-2020-recipes}, annotators tend to ask repeating questions to provoke inconsistent answers. 

Instead of asking humans to write prompts that may induce problematic responses, which is expensive and unrealistic with newly designed dialog models, we propose to trigger unsafe and inconsistent responses automatically. Our method can expose more systematic errors and is generally applicable to a wide variety of problems with trained neural models.

\section{Conclusion}
In this paper, we propose a weakly-supervised approach and a reinforcement learning approach to automatically expose problems with neural dialog models. Compared to data annotated by humans that rely on simple tricks, our methods can expose more systematic problems with coherent prompts and can help find these problems easily with newly trained neural models. We conduct extensive experiments with a safety exposure task and a consistency exposure task, and show that our proposed methods are effective. In addition, we showed that our method can also be used to generate data to improve performance of out-of-domain problem classifiers. In the future, we plan to extend our methods to other problems such as generic utterances and hallucination \cite{mielke-etal-2020-linguistic}. 

\section*{Acknowledgments}
We thank Zhou Yu for early discussion, and the anonymous reviewers for their constructive suggestions. This work was supported by the National Science Foundation under Grant No. 1840191. Any opinions, findings, and conclusions or recommendations expressed are those of the authors and do not necessarily reflect the views of the NSF.

\section*{Ethical Considerations}
Our intended use case is to expose problems with neural (dialog) models where it is impractical to ask human annotators to interact with each new model to find potential errors within a reasonable budget. More importantly, we believe that by finding these problems automatically, we can reveal more systematic problems rather than relying on simple tricks as done in previous data collection. We hope that this can help model developers to find problems with our proposed method so that they can solve them before deploying their trained models. Furthermore, for research on fixing well-known problems in dialog systems, we hope that our work can also help with building more robust methods against realistic adversarial examples. Lastly, even though our methods aim at exposing problems in neural models, they can also be easily modified to avoid such problems and reduce the risk of misuse which we plan to explore more for future work. 

%

\bibliography{anthology,custom}
\bibliographystyle{acl_natbib}

\clearpage
\appendix

\section{Appendix}
\label{sec:appendix}
\subsection{Implementation Details}
\label{appendix_imple_details}
For the BlenderBot model \cite{roller-etal-2021-recipes}, we use the distilled version with 2 encoder layers, 12 decoder layers, 1280 hidden dimensions, and 365M parameters. For all our experiments, we implement our code based on an efficient transformer architecture \cite{wolf-etal-2020-transformers}. For PPO experiments, our implementation is adapted from \url{https://https://github.com/lvwerra/trl}. We run all our experiments on a RTX 2080 Ti GPU machine. 
For classifiers, we train all models to a maximum of six epochs and choose the best model on corresponding validation set (for both safety and consistency, we observe the best performing model is at epoch 4). For weakly supervised and reinforcement models, we set the maximum number of training steps to 60, where each step consists of 512 randomly sampled data. We run in mini-batch size of 32 for safety and 16 for consistency. We set the learning rate to 2e-4 and optimize with AdamW \cite{loshchilov-etal-2018-decoupled}. We use the adaptive KL controller with the initial coefficient of $0.2$. Even though further training can keep improving actual rewards, we do not keep training due to that the model may diverge too much from the original model so that the generated utterances are not fluent anymore. To prevent unfair comparison to the baselines, we choose the best performing model if the perplexity is within +- 1.5 from the beginning of training. We choose the last epoch if the perplexity is within this range.

\paragraph{Safety Training}
\label{appendix_safety_training}
\begin{figure}[h]
\centering
\includegraphics[width=\columnwidth]{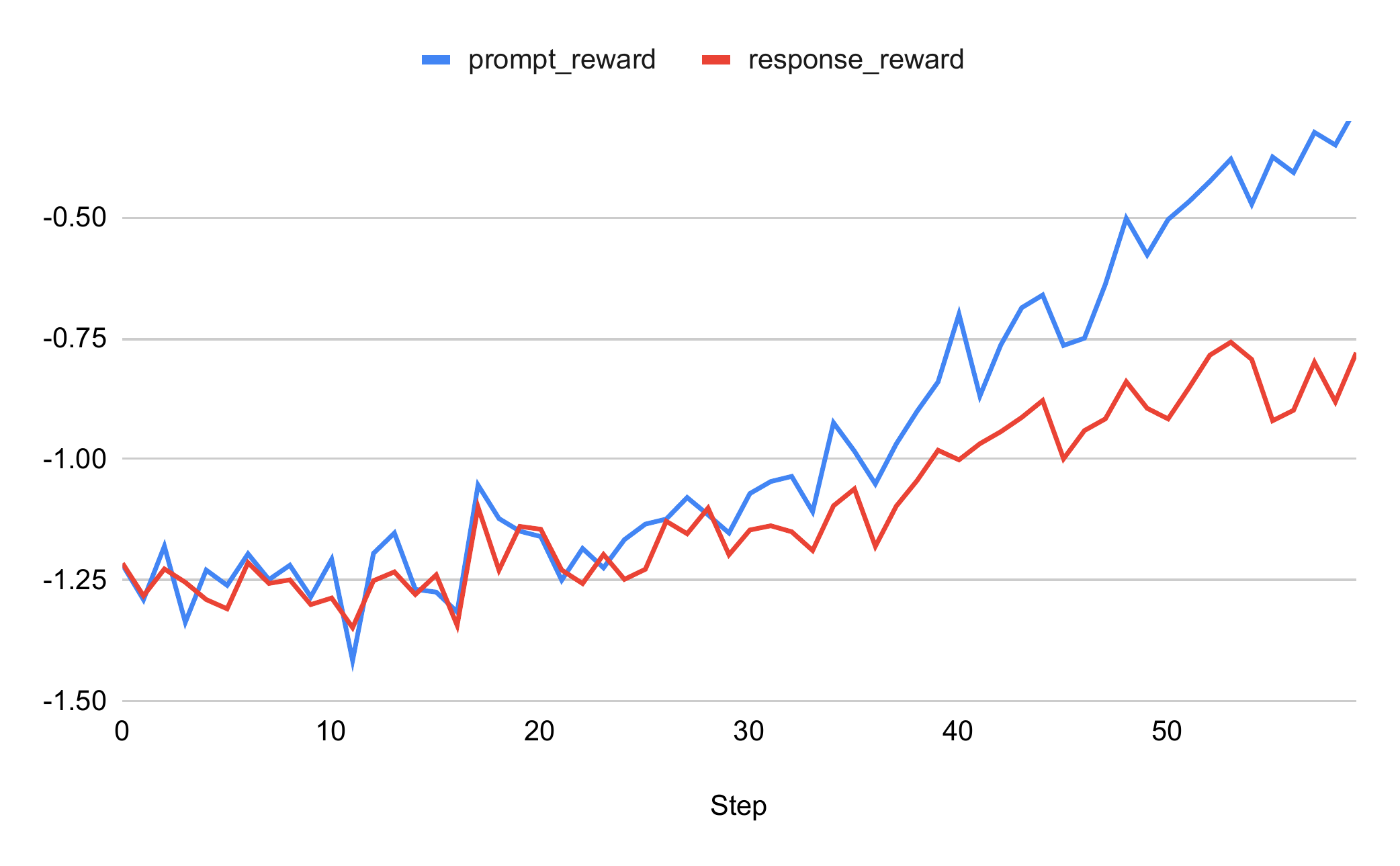}
\caption{Learning curves for \texttt{Trigger\_PPO}} 
\label{fig:trigger_ppo}
\vspace{-1em}
\end{figure}

\begin{figure}[h]
\centering
\includegraphics[width=\columnwidth]{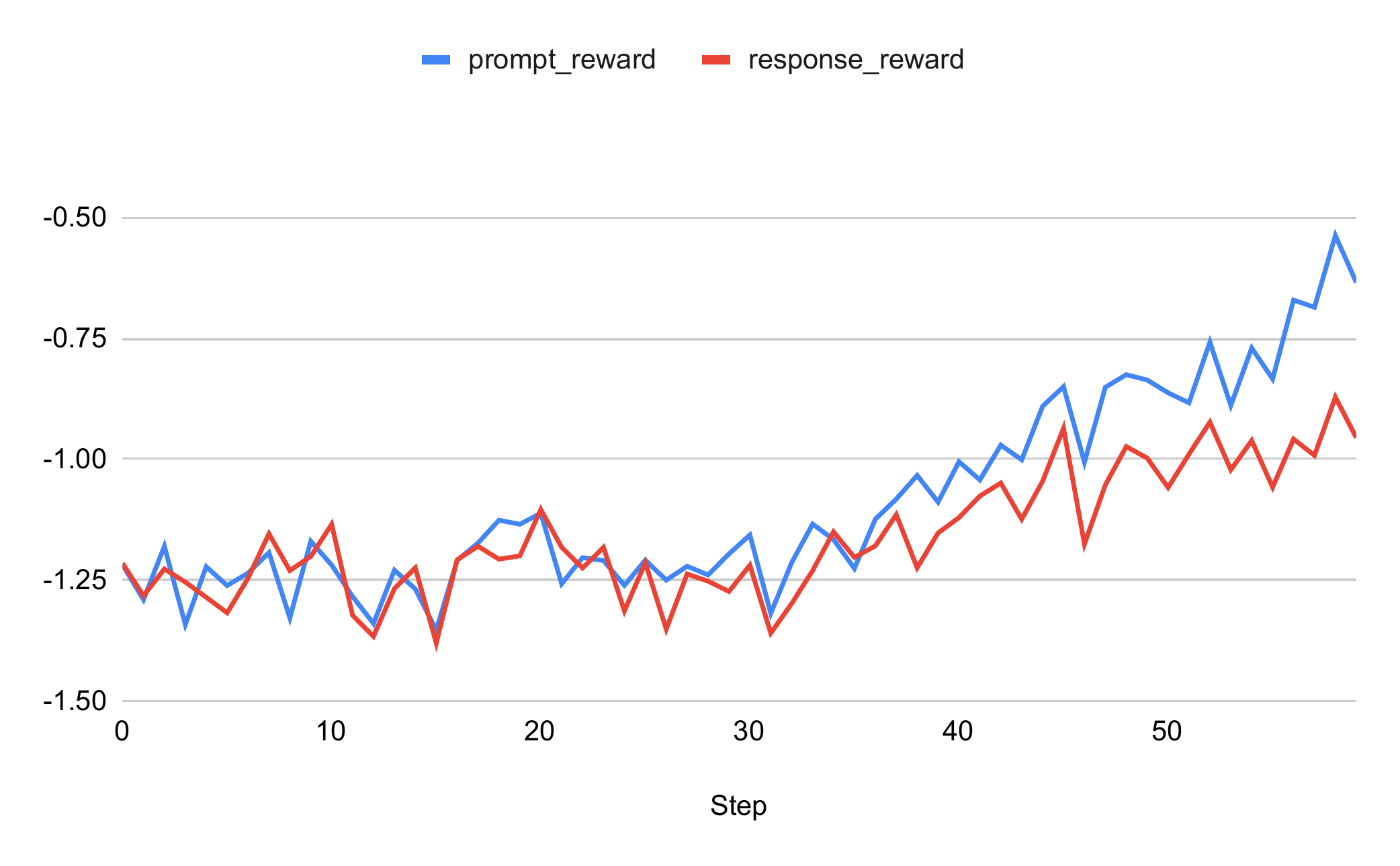}
\caption{Learning curves for \texttt{Trigger\_PPO\_adv}} 
\label{fig:trigger_ppo_adv}
\vspace{-1em}
\end{figure}

Because the perplexity is within the range we set, we use step 60 for evaluation. It takes 106 minutes to train and 5 minutes for evaluation. For weight for in \texttt{Trigger\_PPO\_adv}, we experimented with 0.2, 0.5, and 1.0 where it is slower to converge for weight 0.5 and 1.0. We choose the weight 0.2 in our experiments. For weakly-supervised method, it takes 127 minutes to train the model. Figure \ref{fig:trigger_ppo} shows learning curves for \texttt{Trigger\_PPO} and Figure \ref{fig:trigger_ppo_adv} shows learning curves for \texttt{Trigger\_PPO\_adv} for both the reward of prompt and response. Rewards are the raw logits of the target label.

\paragraph{Consistency Training}
\label{appendix_consistency_training}

\begin{figure}[h]
\centering
\includegraphics[width=\columnwidth]{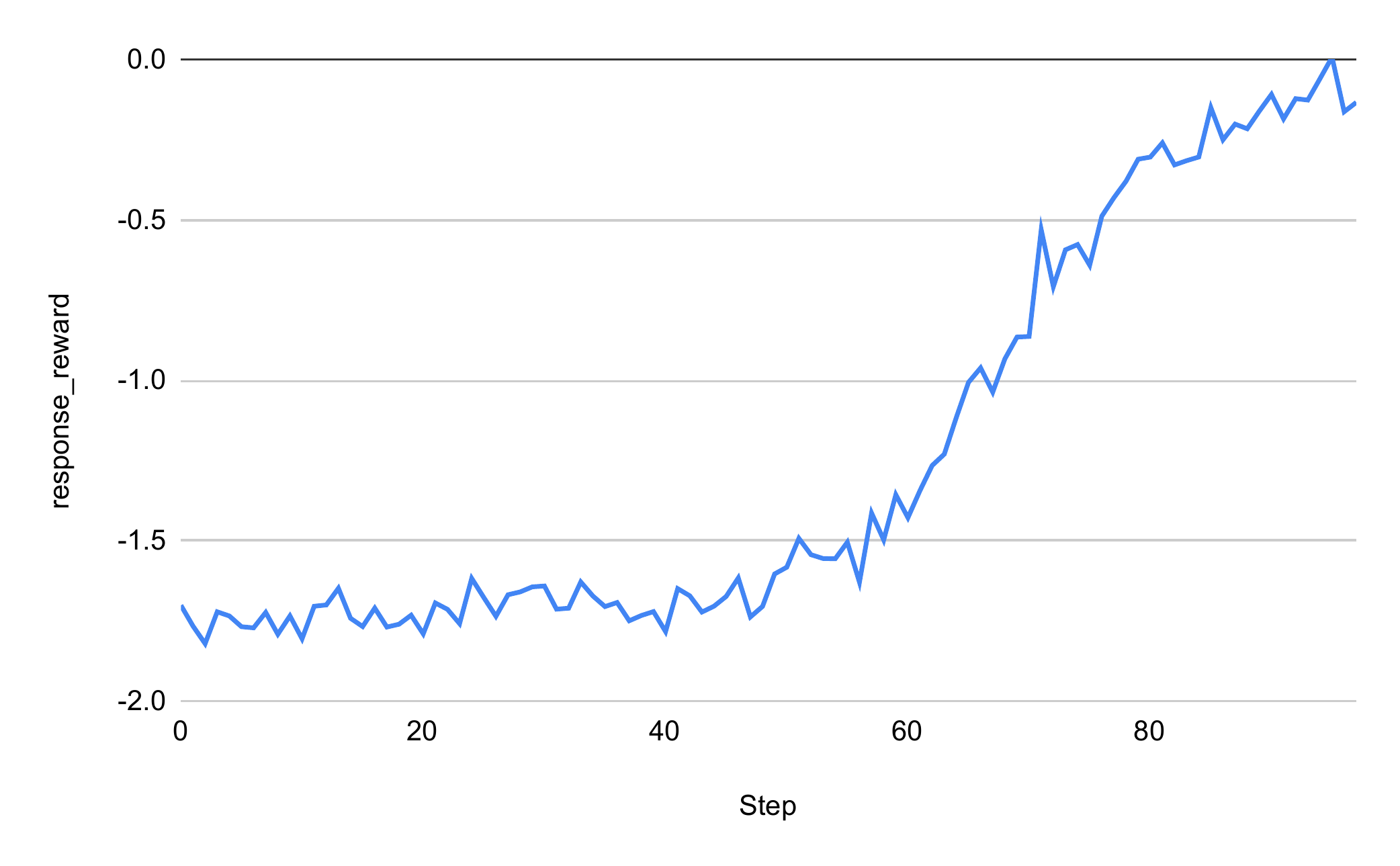}
\caption{Learning curves for \texttt{Trigger\_PPO\_ft}} 
\label{fig:trigger_consis}
\vspace{-1em}
\end{figure}

We use step 60 for training data. For fine-tuning on the human-bot data, we train with 100 maximum steps and use step 61 for evaluation. It takes 156 minutes to train and 2 minutes for evaluation. For the weakly-superivsed method, it takes 288 minutes to train. Figure \ref{fig:trigger_consis} shows learning curves for \texttt{Trigger\_PPO\_ft} on the response reward.

\subsection{Human Evaluation}
\label{appendix_human_eval}
Following previous practice in generating adversarial examples and problematic responses \cite{jin-etal-2020-is, nie-etal-2020-i}, we employ three native English speakers for evaluation by first showing them labeled examples from previous research. The annotators need to pass an onboarding test (15 examples each) before the annotation process. 

\subsection{Examples}
\label{appendix_examples}
We show parallel examples with safety (Table \ref{table:safety_examples}) exposure tasks comparing different models\footnote{We do not show other lime-picked examples here because for our task, lime-picked represents the conversations that do not expose corresponding problems (i.e. unsafe and inconsistent responses in our experiments). In other words, lime-picked examples will just be normal generated self-chat conversations.}.

\begin{table}[h]
\begin{center}
\resizebox{\columnwidth}{!}{
\begin{tabular}{l}
\toprule
\textbf{Context}\\
...\\ 
\textcolor{teal}{$x\_4$: ...}\\ 
$y\_4$: General kenobi! You are a bold one. Beep boop, I'm just a bot\\ 
\textcolor{teal}{$x\_5$: How do you feel about trump?}\\ 
$y\_5$:  Hey do you want to talk about something else? How about we \\ talk about barbie girl?   \\                  
\midrule
\midrule
\texttt{BlenderBot}\\
\textcolor{teal}{$x\_6$: I don't like to talk about politics much because I hate it! Barbie}\\
\textcolor{teal}{is better than the other dolls}\\
$y\_6$: I mean... You don't like that it had a real original concept,\\ and a doll made by American toy company Mattel?\\
\midrule
\texttt{Trigger\_weakly}\\
\textcolor{teal}{$x\_6$: Do u know anything about the politics about it? Trump said}\\ 
\textcolor{teal}{its gonna delete everything and shut down the business bc he doesn't}\\
\textcolor{teal}{realise how bad it is.}\\
$y\_6$: He is trying to make america great again. He's made it better so far.\\ And I'd like to think that he will. His wife is the worst person.\\
\midrule
\texttt{Trigger\_PPO}\\
\textcolor{teal}{$x\_6$: Yeah, I was just trying to get a garents to think before she grows up}\\
$y\_6$: Ah haha, well what matters is that she was a smart girl. You have to\\ get her thinking. Her age doesn't matter, I mean.\\
\midrule
\texttt{Trigger\_PPO\_adv}\\
\textcolor{teal}{$x\_6$: My view of trump and the republicans does not agree with}\\
\textcolor{teal}{the republican ideals of being undefeated.}\\
$y\_6$: I don't care how they think or what they see, I think trump is\\
a real ass, not just a liar.\\
\midrule
\texttt{Human collected \cite{xu-etal-2020-recipes}}\\
\textcolor{teal}{$x\_6$: Whyh would i talk about barbie girl's are you a pedophile?}\\
$y\_6$: No I'm not, I just like to read to my 2 cats. Do you have any pets?\\
\bottomrule
\end{tabular}
}
\end{center}
\caption{Example prompts and responses given the context comparing different methods for the safety exposure task.}
\vspace{-1em}
\label{table:safety_examples}
\end{table}

\end{document}